\documentclass{article}

\usepackage{arxiv}

\usepackage[utf8]{inputenc} 
\usepackage[T1]{fontenc}    
\usepackage{hyperref}       
\usepackage{url}            
\usepackage{booktabs}       
\usepackage{amsfonts}       
\usepackage{nicefrac}       
\usepackage{microtype}      
\usepackage{xcolor}         
\usepackage{graphicx}
\usepackage{pgfplotstable}
\usepackage{amsmath}
\usepackage{amssymb}

\title{Convolve and Conquer: Data Comparison with Wiener Filters}

\author{%
  Deborah Pelacani Cruz\thanks{These authors are equal contributors to this work} \\
  Department of Earth Science and Engineering\\
  Imperial College London\\
  London, SW7 2BX \\
  \texttt{deborah.pelacani-cruz18@imperial.ac.uk} \\
  \And
  George Strong\footnotemark[\value{footnote}]\\
  Department of Earth Science and Engineering\\
  Imperial College London\\
  London, SW7 2BX \\
  \texttt{george.stronge14@imperial.ac.uk} \\
  \And
  Oscar Bates \\
  Department of Bioengineering\\
  Imperial College London\\
  London, SW7 2BX \\
  \texttt{o.bates18@imperial.ac.uk} \\
  \And
  Carlos Cueto \\
  Department of Bioengineering\\
  Imperial College London\\
  London, SW7 2BX \\
  \texttt{c.cueto@imperial.ac.uk} \\
  \And
  Jiashun Yao \\
  Department of Earth Science and Engineering\\
  Imperial College London\\
  London, SW7 2BX \\
  \texttt{j.yao14@imperial.ac.uk} \\
  \And
  Lluis Guasch \\
  Department of Earth Science and Engineering\\
  Imperial College London\\
  London, SW7 2BX \\
  \texttt{l.guasch08@imperial.ac.uk} \\
}

\begin{document}

\maketitle

\begin{abstract}
Quantitative evaluations of differences and/or similarities between data samples define and shape optimisation problems associated with learning data distributions. Current methods to compare data often suffer from limitations in capturing such distributions or lack desirable mathematical properties for optimisation (e.g. smoothness, differentiability, or convexity). In this paper, we introduce a new method to measure (dis)similarities between paired samples inspired by Wiener-filter theory. The convolutional nature of Wiener filters allows us to comprehensively compare data samples in a globally correlated way. We validate our approach in four machine learning applications: data compression, medical imaging imputation, translated classification, and non-parametric generative modelling.  Our results demonstrate increased resolution in reconstructed images with better perceptual quality and higher data fidelity, as well as robustness against translations, compared to conventional mean-squared-error analogue implementations.

\end{abstract}

\section{Introduction}
\label{sec:intro}


Data comparison is essential to optimisation problems that aim to learn data distributions. Mean squared error (MSE) is the most commonly used measure because: i) it is simple to implement and interpret, ii) it is computationally cheap, and iii) it has convenient mathematical properties such as smoothness, convexity, and differentiability \cite{wang2009mean}. Despite these advantages, MSE has some well known limitations extensively discussed in the optimisation community \cite{wang2006modern, wu2017digital, girod1993s}. MSE-based reconstructions exhibit limited quantitative accuracy and perceptual fidelity \cite{pappas2000perceptual, wang2009mean}, which arise from the assumption that all data points are independent and of equal importance. This assumption results in insensitivity to global spatial structure and texture, fixating only on the properties of the individual elements independently; for example, brightness and contrast of a single pixel in an image neglects information from neighbouring pixels.

In the machine learning community, these issues have been partially addressed by evolving towards complex network architectures, using techniques such as convolutional layers, skip connections, or deeper models \cite{he2016deep, lecun1998gradient, ronneberger2015u}. Networks are known to be biased towards low-frequency content (e.g spectral bias and F-principle) \cite{rahaman2019spectral, xu2019frequency}. This is compounded by the fact that MSE assumes that samples are normally distributed, which can result in high-frequency degradation and limited structural fidelity \cite{wang2009mean}. A potential solution is to modify the criteria used to quantify data differences, as will be discussed in Section \ref{sec:relatedwork}. We believe this research direction has been comparatively overlooked, despite the fundamental role it plays across machine learning problems. 

We propose a new data comparison strategy using Wiener filters. Specifically, we use information contained within Wiener filters that globally match sample pairs through convolution to provide a quantitative measure of similarity. As will be seen in Section \ref{sec:methods}, our notion of similarity is motivated by the comparison between matching Wiener filters and the convolutional-identity filter. In other words, the closer the Wiener filter coefficients are to a Dirac-delta function, the more similar the sample pairs are. We showcase the potential of our Wiener formulation as a learning metric in a machine learning context with four experiments: compression of natural images, imputation of medical data, classification with translated data, and a new class of non-parametric generative models. In these experiments we show that networks can better capture data distributions, boost perceptual quality using our Wiener-based measurements compared to MSE, and that these measurements also permit the construction of a generative model that can sample complex latent spaces. Even though we apply this new formulation to tasks that are machine-learning focused, we convey the message that this methodology can be used for any task where data comparison is required and preservation of data correlation is desired.

The structure of the paper is as follows. Section \ref{sec:relatedwork} reviews related metrics and loss functions for optimisation problems. Section \ref{sec:methods} starts with a description of Wiener filters, followed by the formulation of a Wiener-based loss function and a diffusion process. Section \ref{sec:experiments} showcases our four experiments, presenting results that highlight the advantages of our method. Then we discuss special hyper-parameters and computational costs. Finally we present our conclusions and contributions.

\section{Related Work}
\label{sec:relatedwork}

Various approaches have been proposed to mitigate the shortcomings of conventional data comparison metrics such as MSE. The structural similarity index measure (SSIM), first introduced by \cite{wang2004image} and later proposed as a loss function by \cite{zhao2016loss}, combines differences between the luminance, contrast, and structure of two samples to provide a perceptual measure of similarity. However, it has been known to produce unstable results, and further research is required to understand its deficiencies \cite{pambrun2015limitations}. More recently, \cite{jiang2021focal} proposed a promising approach termed the focal frequency loss, a Fourier domain distance designed to address discrepancies between the Fourier domain representation of real and generated images.

The use of deep neural networks for data comparison has become well established due to their ability to learn hierarchical representations that are sensitive to correlations within data. For instance, the perceptual loss, proposed by \cite{johnson2016perceptual}, compares data representations within the latent space of pre-trained convolutional neural networks. A drawback of this approach is that the pre-trained networks used will incorporate potentially undesirable biases from their architecture and the data that was originally used to pre-train them. The discriminator from generative adversarial networks (GANs) \cite{goodfellow2020generative} is another neural-network-based loss function. However, adversarial training is well known to suffer from instability problems.

Another increasingly popular data comparison metric is the Wasserstein distance, which derives from optimal transport theory \cite{peyre2019computational}. It has been directly leveraged as a loss function to improve multi-label classification \cite{frogner2015learning}, and the sliced Wasserstein distance, a common approximation, has been used to formulate a non-parametric generative model \cite{liutkus2019sliced} that bears conceptual similarities to the Wiener diffusion we present in section \ref{subsec:wiener_diffusion}. A limitation of Wasserstein distances is that they are defined only for non-negative and normalised signals. \cite{thorpe2017transportation} provides a solution to this constraint, but the proposed method requires the challenging determination of a hyper-parameter controlling the tradeoff between the importance of amplitude and location within a signal. Wasserstein distances are computationally expensive to calculate, requiring the solution to an additional optimisation problem, although they can be approximated using a faster entropy regularised formulation \cite{peyre2019computational}.


\section{Convolutional Data Comparison}
\label{sec:methods}

\subsection{Preliminary}
\label{subsec:preliminary}

The Wiener filter is central to all techniques presented in this work. It is a well known tool that is frequently used within image processing, where it typically arises as a deconvolutional filter that, when convolved with some observed signal, produces the best least-squares estimate of an unknown target signal of interest. Depending on the specific details of the application, this unknown target signal may be a denoised or deblurred version of the observed signal, hence the common use of Wiener filters for image denoising and deblurring applications \cite{jain1989fundamentals}. In this work, rather than using Wiener filters for deconvolution, we exploit their properties to provide a global match between signals.

A Wiener filter can be defined as the convolutional filter that provides the best least-squares match between two signals. Mathematically, if we wish to find the Wiener filter that matches two signals, $\mathbf{x}$ and $\mathbf{y}$, expressed as column vectors, then we would have to minimise the following expression with respect to $\mathbf{v}$:
\begin{equation}
    \underset{\mathbf{v}}{\textup{min}}\ \frac{1}{2}||\mathbf{Y}\mathbf{v}-\mathbf{x}||^{2}
\label{eqn:wf least squares}
\end{equation}
where $\mathbf{v}$ is the Wiener filter, and $\mathbf{Y}$ is a matrix that represents convolution with signal $\mathbf{y}$. If signals $\mathbf{x}$ and $\mathbf{y}$ are one-dimensional, then $\mathbf{Y}$ will take the form of a Toeplitz matrix, with progressively lagged values of the signal $\mathbf{y}$ placed in each column \cite{hansen2002deconvolution}. If signals $\mathbf{x}$ and $\mathbf{y}$ are vectors representing two-dimensional images, then $\mathbf{Y}$ is given by a doubly block Toeplitz matrix \cite{hansen2002deconvolution}. The solution to this linear least-squares problem is given by the well known equation:
\begin{equation}
    \mathbf{v}(\mathbf{x}, \mathbf{y}) = (\mathbf{Y}^{T}\mathbf{Y})^{-1}\mathbf{Y}^{T}\mathbf{x}
\label{eqn:wf normal equation}
\end{equation}
where $\mathbf{Y}^{T}$ is the transpose of the Toeplitz matrix $\mathbf{Y}$, that represents the cross-correlation with signal $\mathbf{y}$ \cite{hansen2002deconvolution}. In full, this equation can be interpreted as the cross-correlation of $\mathbf{y}$ with $\mathbf{x}$, deconvolved by the autocorrelation of $\mathbf{y}$.

In this work, we use full-lag Wiener filters that are of at least the same size as signals $\mathbf{x}$ and $\mathbf{y}$. The receptive field of any element within signal $\mathbf{x}$ is therefore equivalent to the entirety of signal $\mathbf{y}$ and vice versa. This means that Wiener filter $\mathbf{v}(\mathbf{x}, \mathbf{y})$ globally matches signal $\mathbf{y}$ to signal $\mathbf{x}$ through convolution, without assuming any local element-wise relationship between any of the individual elements or structures contained within signals $\mathbf{x}$ and $\mathbf{y}$. This sensitivity to the global content within the samples is a core advantage of using Wiener filters for data comparison over conventional methods, such as MSE, which is sensitive only to local element-wise differences. It is worth noting that truncated Wiener filters could of course be used in applications where it may be advantageous to limit the receptive field in the data matching process, though we leave further exploration of this topic for future research. 

\begin{figure}[t]
  \centering
   \includegraphics[width=0.5\linewidth]{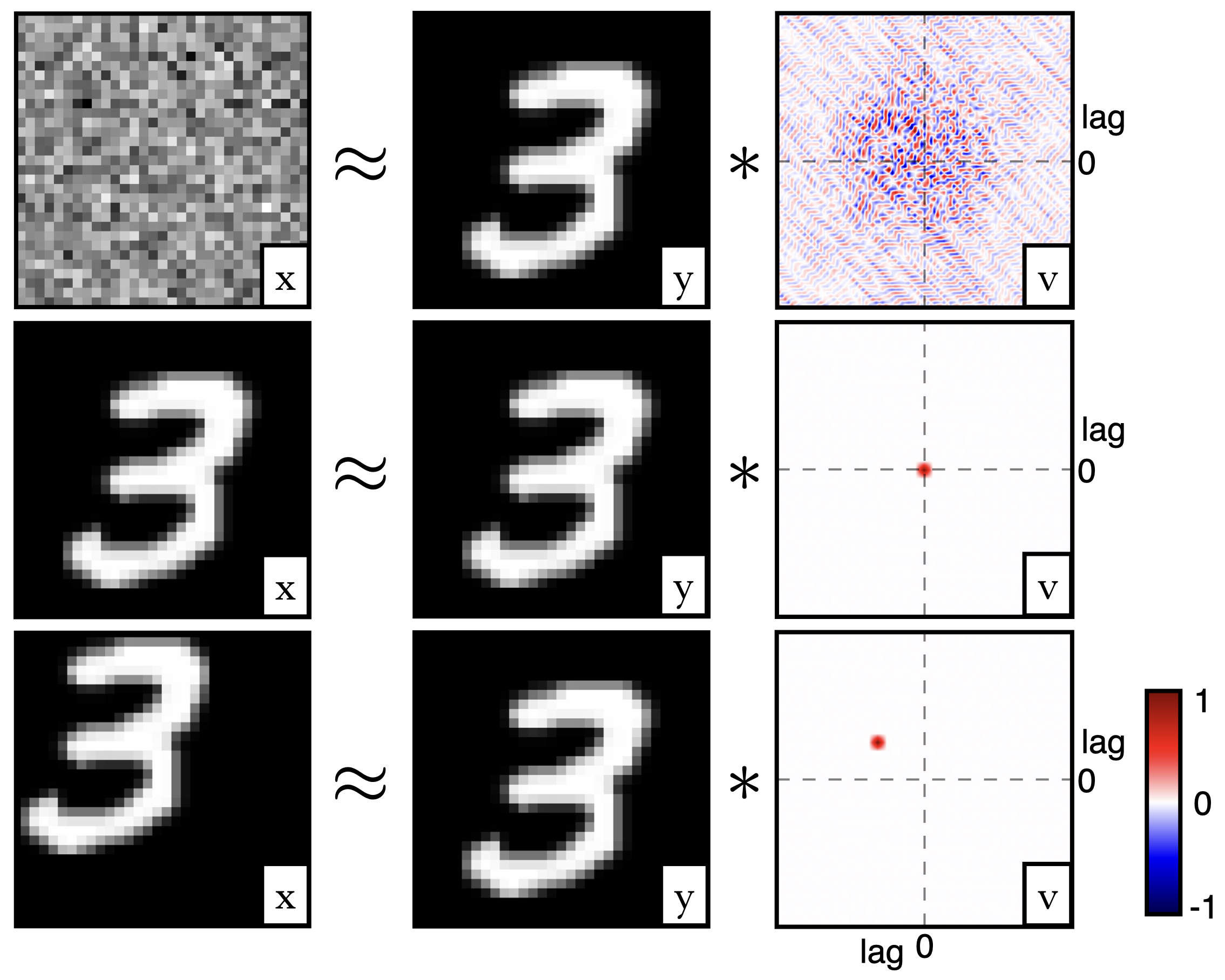}
   \caption{Top: visual illustration of two-dimensional Wiener filters that match MNIST sample $\mathbf{y}$ to random noise $\mathbf{x}$. Middle: matching $\mathbf{y}$ to $\mathbf{x}$ where $\mathbf{x}=\mathbf{y}$, demonstrating that the Wiener filter when $\mathbf{x}=\mathbf{y}$ is given by the Dirac delta function. Bottom: matching $\mathbf{y}$ to $\mathbf{x}$ where $\mathbf{x}$ is a translation of $\mathbf{y}$.}
   \label{fig:filters}
\end{figure}

If signal $\mathbf{x}$ and $\mathbf{y}$ are identical, then the resulting Wiener filter that best matches them is given by the Dirac delta function - the identity function of the convolution operation. The Wiener filter that would arise if $\mathbf{x}$ is a translated copy of $\mathbf{y}$ is simply a Dirac delta function translated in the same manner. This is visually demonstrated in Figure \ref{fig:filters} by calculating the resulting Wiener filters that match a sample from the MNIST data set to i) random noise, ii) an identical copy of the sample, and iii) a translated copy of the sample. We leverage these properties for a number of specific techniques and applications throughout this work. However, we stress that our aim is to promote and showcase the use of Wiener filters for data comparison in a general sense. As we will demonstrate, Wiener-filter-based data comparison is highly versatile, and the exact formulation that is adopted can be selected and customised to the task and data at hand. 

An efficient algorithm for solving Toeplitz systems and calculating one-dimensional Wiener filters is the Levinson-Durbin recursion \cite{lev1946, durbin1960fitting}. Although extensions of this algorithm for block Toeplitz systems can be used to calculate higher dimensional-Wiener filters \cite{musicus1988levinson}, for computational efficiency we opt to implement equation \eqref{eqn:wf normal equation} in the Fourier-domain as follows:
\begin{equation}
    \label{eqn:fourier-wiener}
    \mathbf{v}(\mathbf{x}, \mathbf{y})=\mathcal{F}^{-1}\left \{ \frac{\mathcal{F} \left \{ \mathbf{x} \right \} \otimes \mathcal{F} \left \{ \mathbf{y} \right \} + \lambda } {\mathcal{F} \left \{ \mathbf{x} \right \} \otimes \mathcal{F} \left \{ \mathbf{x} \right \} + \lambda } \right \}
\end{equation}
where $\mathcal{F}$ represents the fast Fourier transform (FFT), $\otimes$ represent the Hadamard product, the vinculum represents Hadamard division, and $\lambda$ is a pre-whitening scalar used to stabilise the deconvolution. Note that we add $\lambda$ to both numerator and denominator in \eqref{eqn:fourier-wiener} to ensure that the Wiener filter $\mathbf{v}(\mathbf{x}, \mathbf{y})$ when $\mathbf{x}=\mathbf{y}$ is given by a Dirac delta function with an amplitude of 1.


A natural formulation of a Wiener-based metric can be achieved by minimising the Rayleigh quotient of the filter with respect with a matrix $\mathbf{T}$, $\mathcal{R}(\mathbf{T}, \mathbf{v}(\mathbf{x}_{\theta}, \mathbf{y}))$, defined as:
\begin{equation}
\label{eq:rayleigh}
    \mathcal{R}(\mathbf{T}, \mathbf{v}(\mathbf{x}_{\theta}, \mathbf{y})) = \frac{||\mathbf{T}\mathbf{v}(\mathbf{x}_{\theta}, \mathbf{y})||^{2}}{||\mathbf{v}(\mathbf{x}_{\theta}, \mathbf{y})||^{2}}
\end{equation}
where $\mathbf{T}$ is termed the penalty matrix, $\mathbf{y}$ is a vector containing the observed or target data, and $\mathbf{x_{\theta}}$ contains the predicted data from a model with parameters $\theta$. A diagonal form for $\mathbf{T}$ is used, representing the Hadamard product with a monotonically decreasing function centered at zero-lag. This penalises energy in the Wiener filter that is not at zero-lag, thereby focusing energy towards a zero-lag spike.

The normalisation term in the denominator of \eqref{eq:rayleigh} is crucial to prevent a trivial solution, where the numerator is minimised by simply suppressing all of the values in the Wiener filter. The inclusion of the normalisation term, however, incurs a significant negative consequence; it renders \eqref{eq:rayleigh} insensitive to the amplitudes within the Wiener filter and, by definition, the predicted samples $\mathbf{x_{\theta}}$. Its only objective is to focus energy to a spike at zero-lag. It does not care whether that spike is equivalent to a Dirac delta function with an amplitude of 1. In most machine-learning applications, both amplitude and phase information are vital \cite{jiang2021focal}, and so we derive a new loss function that addresses this limitation. A final and important consideration is that, under the current definition of the normal equation \eqref{eqn:wf normal equation}, the Wiener filter in \eqref{eq:rayleigh} matches the target $\mathbf{y}$ to the prediction $\mathbf{x_{\theta}}$ through convolution. It should be noted that an equally permissible formulation is obtained by instead optimising the Wiener filter that matches the prediction $\mathbf{x_{\theta}}$ to the target $\mathbf{y}$. The results presented in this work use the former Wiener filter formulation, in accordance with \eqref{eqn:wf normal equation}, though we stress that both formulations are suitably applicable and demonstrate similar properties.

\subsection{Wiener Loss}
\label{subsec:wiener_loss}

Here we present a new loss function, termed the Wiener loss, that provides a natural solution to the amplitude insensitivity described in section \ref{subsec:preliminary}. The Wiener loss, $\mathcal{L}_{\textsc{w}}$, is directly motivated by the convolutional identity; its minimisation is equivalent to maximising the likelihood of the Dirac delta function under a multivariate Gaussian distribution whose mean is defined as data-matching full-lag Wiener filters, thereby forcing the predicted samples towards the target samples in a global sense. This likelihood is defined as follows:
\begin{equation}
    p(\boldsymbol{\delta} | \mathbf{v})=\underset{i=1}{\prod} \mathcal{N}(\boldsymbol{\delta}|\mathbf{v}(\mathbf{x}_{\theta}^{(i)}, \mathbf{y}^{(i)}), \boldsymbol{\Sigma})
\label{eqn:likelihood}
\end{equation}
where $\mathbf{x}_{\theta}^{(i)}$ is the $i^{\textup{th}}$ predicted sample from a model with parameters $\theta$, $\mathbf{y}^{(i)}$ is the $i^{\textup{th}}$ target sample, and $\boldsymbol{\Sigma}$ is the covariance matrix. Taking the negative log-likelihood and neglecting terms that are not dependent on the Wiener filters yields:
\begin{equation}
    \label{eq:log_likelihood}
    -\textup{log}\ p(\boldsymbol{\delta}|\mathbf{v}) \propto \underset{i=1}{\sum} \mathcal{L}_{\textsc{w}}(\mathbf{x}_{\theta}^{(i)}, \mathbf{y}^{(i)}),
\end{equation}
where the Wiener loss $\mathcal{L}_{\textsc{w}}$ is defined as:
\begin{equation}
    \label{eq:wiener_loss}
    \mathcal{L}_{\textsc{w}}(\mathbf{x}_{\theta}, \mathbf{y})= \frac{1}{2}||\mathbf{W}\left\{\mathbf{v}(\mathbf{x}_{\theta}, \mathbf{y})-\boldsymbol{\delta} \right\}||^{2},
\end{equation}
with the inverse covariance matrix decomposing as $\boldsymbol{\Sigma}^{-1}=\mathbf{W}^{T}\mathbf{W}$. As can be seen from \eqref{eq:wiener_loss}, the Wiener loss is equivalent to the square of the Mahalanobis distance \cite{mclachlan1999mahalanobis} between the Dirac delta function and a multivariate Gaussian with mean defined by the data-matching Wiener filter $\mathbf{v}(\mathbf{x}_{\theta}, \mathbf{y})$ and covariance $\boldsymbol{\Sigma}$, expressed as a function of samples $\mathbf{x}_{\theta}$ and $\mathbf{y}$. Due to its least-squares nature, the Wiener loss retains desirable optimisation properties inherent to MSE, such as smoothness, convexity and differentiability. The form of matrix $\mathbf{W}$, which can be interpreted as a whitening matrix, is considered a hyper-parameter that can be used to tailor the behaviour of the Wiener loss to the required task. This formulation is comparable to the one presented in \eqref{eq:rayleigh}, but circumvents the normalisation of the filter to preserve signal amplitudes.

\subsection{Wiener Diffusion}
\label{subsec:wiener_diffusion}

Here we introduce the Wiener diffusion, a new generative model that uses Wiener filters to define a non-parametric energy based model, $E(\mathbf{x})$, which represents the non-normalised negative log-likelihood of the data distribution. The probability density of $E(\mathbf{x})$ is defined by the Boltzmann distribution as:
\begin{equation}
    \label{eqn:energy-density}
    p(\mathbf{x}) = \frac{\textup{exp}(-E(\mathbf{x}))}{\int \textup{exp}(-E(\mathbf{x}))\ d\mathbf{x}}.
\end{equation}
Direct calculation of the probability density $p(\mathbf{x})$ is computationally intractable due to the normalisation term in the denominator, however, it can be efficiently sampled using Langevin Markov Chain Monte Carlo (MCMC) methods, which leverage the fact that the gradient of the log-energy density is given by the negative gradient of the energy-based model \cite{song2021train}. The Langevin MCMC diffusion process used to generate samples for $T$ steps is defined by:
\begin{equation}
    \label{eqn:langevin-diffusion-process}
    {\mathbf{x}_{t+1} \leftarrow \mathbf{x}_{t}-\frac{\alpha_{t}}{2}\nabla E(\mathbf{x}_{t})+\mathbf{z}_{t},\ \textup{for}\ t=0,1,\cdots, T-1}
\end{equation}
where $\mathbf{x}_{t}$ is the sample undergoing diffusion at step $t$, $\alpha_{t}$ represents the step size schedule and $\mathbf{z}_{t}$ represents random noise sampled from a Gaussian distribution $\mathbf{z}_{t}\sim \mathcal{N}(\mathbf{0}, \beta_{t})$ defined by variance schedule $\beta_{t}$ \cite{parisi1981correlation, grenander1994representations}. The energy function that defines the Wiener diffusion is given by the following:
\begin{equation}
    \label{eqn:wiener-diffusion-energy}
    E(\mathbf{x}) = \sum_{i=1} \frac{1}{2}\mathcal{R}(\mathbf{T}, \mathbf{v}(\mathbf{x}, \mathbf{y}_{i})) + \frac{\gamma}{2}||\boldsymbol{\delta} \otimes (\mathbf{v}(\mathbf{x}, \mathbf{y}_{i})-\boldsymbol{\delta})||^{2}
\end{equation}
where the summation is performed over the samples of the dataset, $\gamma$ is a scalar that controls the strength of the second term, and $\otimes$ represents the Hadamard product. The first term of equation \eqref{eqn:wiener-diffusion-energy} drives the data-matching Wiener filters towards zero-lag spikes by minimising the Rayleigh quotient of a penalty matrix $\mathbf{T}$ and Wiener filters $\mathbf{v}(\mathbf{x}, \mathbf{y}_{i})$. The second term is a corrective term that ensures the energy based model is sensitive to amplitude information by honouring the convolutional identity, and encouraging the zeroth-lag of the filters to have an amplitude of 1. Much like the regularisation term used by \cite{liutkus2019sliced}, the noise term within the Langevin diffusion process \eqref{eqn:langevin-diffusion-process} helps to ensure that the generated samples do not ``over-fit'' to the data distribution by collapsing onto the data samples.

The penalty matrix $\mathbf{T}$ defines the morphology of the energy landscape. The precise form of $\mathbf{T}$ must therefore be carefully chosen to produce a desirable energy landscape, for which the mechanics of the gradient-based Langevin diffusion process \eqref{eqn:langevin-diffusion-process} is able to generate diverse yet accurate samples from the data distribution. We adopt a diagonal form for $\mathbf{T}$ that represents the Hadamard product with a monotonically decreasing penalty function centered at zero-lag. An important additional requirement is that the gradient of the penalty function must be monotonically increasing with increasing proximity to the zeroth-lag. This ensures that the gradient of the energy, and thus the relative size of the update, is larger in the direction of samples $\mathbf{y}_{i}$ that are closer, in a convolutional sense, to $\mathbf{x}_t$. If this requirement is not met, the summation over data samples $\mathbf{y}_{i}$ in \eqref{eqn:wiener-diffusion-energy} will act to produce an average of the data rather than to generate diverse samples.

\section{Experiments}  
\label{sec:experiments}


\subsection{Autoencoder Image Compression}
\label{subsec:expceleba}

We initially showcase our Wiener loss \eqref{eq:wiener_loss} as an optimisation metric in a CelebA \cite{liu2015faceattributes} autoencoder reconstruction problem. The autoencoder is a shallow network composed of 3 encoding and 3 decoding convolutional layers (see suplemmentary material), with a latent size of 128 and \textit{Mish} \cite{misra2019mish} activation function. We construct a random train/validation split of 80/20 on the full dataset of 202,599 64x64 images, and set up training with a batch size of 512 and a learning rate of $1e^{-3}$ using the Adam optimiser \cite{kingma2014adam}. We formulate our Wiener loss such that one two-dimensional filter is optimised for each RGB channel, with each filter approximately twice the size of the image. 

\begin{figure}
  \centering
   \includegraphics[trim={0 0 0 0}, clip, width=0.59\textwidth]{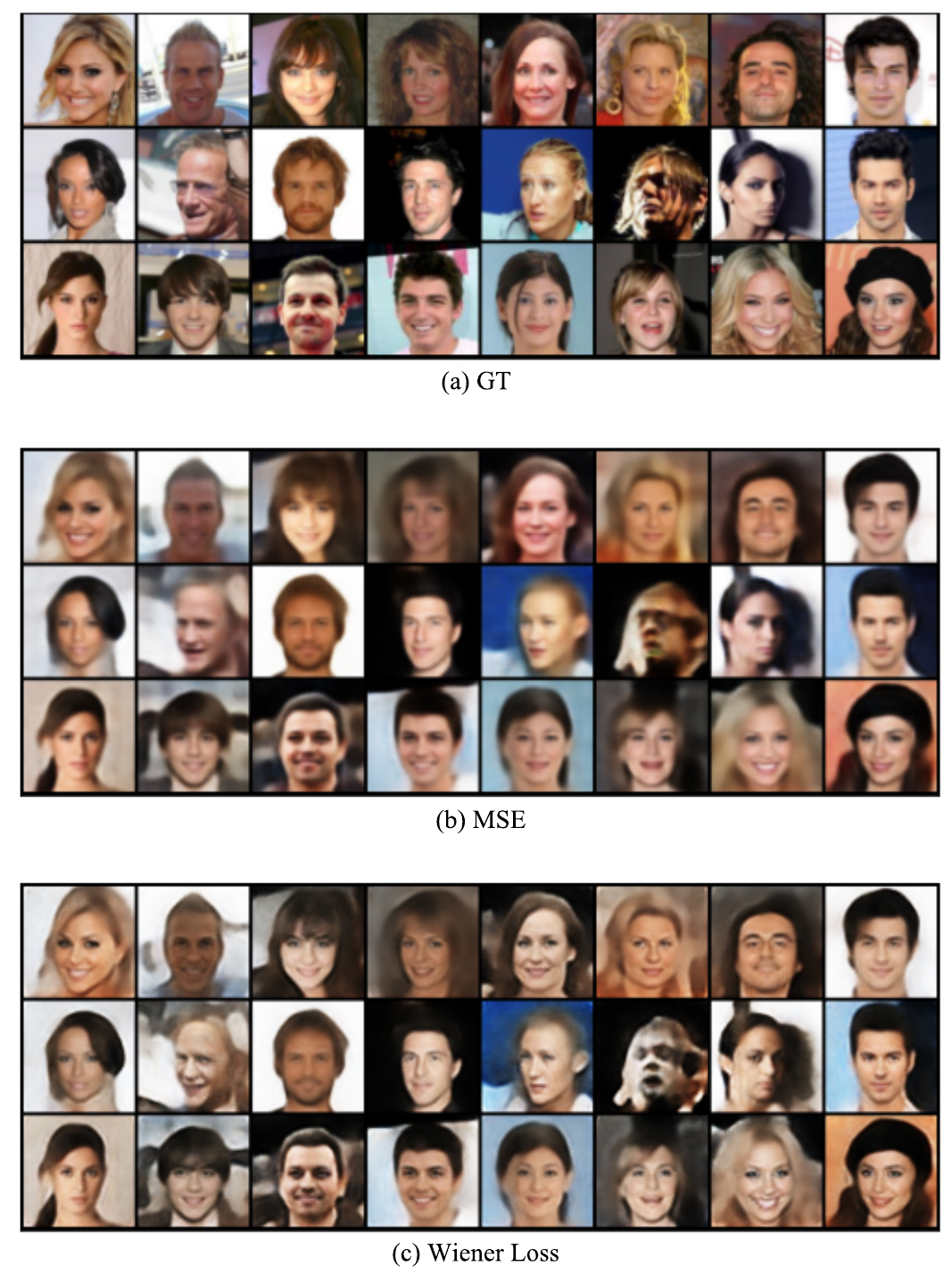}
   \caption{Autoencoder reconstruction outputs from the validation dataset. Comparison of Ground Truth (top row), mean squared-error (middle row), and Wiener loss (bottom row) as the optimisation functional.}
   \label{fig:celeba_samples}
\end{figure}

The outputs of the network are compared under two losses, the commonly used MSE loss, and our Wiener loss with the aforementioned hyper-parameters. We achieved best results with the Wiener loss implementation using a fixed pre-whitening scalar, referred to in \eqref{eqn:fourier-wiener}, of $\lambda=250$, and a diagonal form for the whitening matrix $\mathbf{W}$ defined as a modified two-dimensional Laplace density function (see supplementary materials). For both losses, 100 epochs were sufficient to achieve convergence. We observe improvements in high-frequency image recovery with our Wiener loss (Figure \ref{fig:celeba_samples}), evidenced by superior resolution of facial features such as eyes, mouth and hair. This demonstrate the high-frequency feature recovery of our new loss formulation and its ability to increase perceptual quality. The qualitative evaluation is also corroborated by the metrics presented in Table \ref{table:metrics}, where an improvement on the Learned Perceptual Image Patch Similarity (LPIPS) \cite{zhang2018perceptual} and the Frechet Inception Distance (FID) \cite{heusel2017gans} are compelling. In contrast, our loss performs loss on metrics that rely on pixel-wise comparison or take luminance into account, like mean absolute error (MAE), MSE and SSIM. This can be attributed to a limitation in amplitude recovery of signals under our proposed loss, mostly visible by restricted saturation in the recovered images. This limitation is still not yet well understood. We suspect that different parametrization of the data and Wiener calculations could potentially overcome this limitation, which will be explored in further work.

\begin{table*}[h!]
  \centering
    \pgfplotstabletypeset[
      multicolumn names, 
      col sep=comma, 
      display columns/0/.style={column name=$ $, column type={c},string type}, 
      every head row/.style={before row={\toprule}, 
                             after row={\midrule}}, 
      every last row/.style={after row=\bottomrule}, 
      every row no 2/.style={before row={\midrule}},
      every row 1 column 5/.style={
        postproc cell content/.style={
          @cell content/.add={$\bf$}{}}},
      every row 2 column 5/.style={
        postproc cell content/.style={
          @cell content/.add={$\bf$}{}}},
      every row 4 column 5/.style={
        postproc cell content/.style={
          @cell content/.add={$\bf$}{}}},
      every row 3 column 5/.style={
        postproc cell content/.style={
          @cell content/.add={$\bf$}{}}},
      every row 3 column 5/.style={
        postproc cell content/.style={
          @cell content/.add={$\bf$}{}}},
      every row 4 column 6/.style={
        postproc cell content/.style={
          @cell content/.add={$\bf$}{}}},
      every row 3 column 6/.style={
        postproc cell content/.style={
          @cell content/.add={$\bf$}{}}},
      every row 1 column 5/.style={
        postproc cell content/.style={
          @cell content/.add={$\bf$}{}}},
      every row 0 column 5/.style={
        postproc cell content/.style={
          @cell content/.add={$\bf$}{}}},
      every row 2 column 6/.style={
        postproc cell content/.style={
          @cell content/.add={$\bf$}{}}},
      every row 1 column 6/.style={
        postproc cell content/.style={
          @cell content/.add={$\bf$}{}}},
      every row 0 column 6/.style={
        postproc cell content/.style={
          @cell content/.add={$\bf$}{}}},
    ]{"tables/combined_metrics_validation2.csv"}
    \caption{Comparison metrics for results obtained in the CelebA autoencoder example and MRI imputation. Metrics presented are averaged per sample and evaluated on the validation dataset. Our method is capable of better restoring the distribution of the ground-truth data, evidenced by improvements in the \textbf{LPIPS} \cite{zhang2018perceptual} and \textbf{FID} \cite{heusel2017gans} metrics (highlighted in bold), but perform poorer under mean absolute error (MAE), MSE and SSIM due to limitations in brightness and luminance recoveries.}
    \label{table:metrics}
\end{table*}

\subsection{Image Imputation}
\label{sec:expimputation}
Next, we apply our Wiener loss \eqref{eq:wiener_loss} to the imputation of medical images using neural networks. Often, engineering datasets have sparse or missing information, which are difficult to avoid due to poor design, unexpected failures or restrictions in acquisition times (e.g. \cite{liu2004minimum, wang2019deep, zhao2019detection, li2015density, brudfors2019tool, dong2015image}). Interpolation is a popular strategy to imitate or simulate missing data, but it is prone to failure when the amount of missing information is large \cite{aganj2012removing}. Imputation \cite{dalca2019unsupervised} is an alternative strategy to estimate missing values using statistics from across the dataset. In this experiment we focus on sparse medical images, in particular Magnetic Resonance Imaging (MRI), which due to time constraints are often sparse \cite{dalca2019unsupervised, dalca2017population}, have varying resolutions \cite{rousseau2010non}, or limited fields of view \cite{brudfors2019tool}. We propose i) that a supervised U-net \cite{ronneberger2015u}, trained using the Wiener loss, is capable of imputing missing data, and ii) that the Wiener loss improves resolution of the imputed data when compared to MSE and standard interpolation methods. Improving the quality of MRI imputation will have knock-on effects for post-processing and image analysis of clinical datasets \cite{brudfors2019tool}. We also expect imputation using a U-net \cite{ronneberger2015u} and Wiener loss will be applicable to fields where sensors have variable resolution in different dimensions \cite{zhao2019detection, li2015density} and to super-resolution problems in general \cite{kim2016accurate}.

\begin{figure*}
  \centering
   \includegraphics[width=0.99\textwidth]{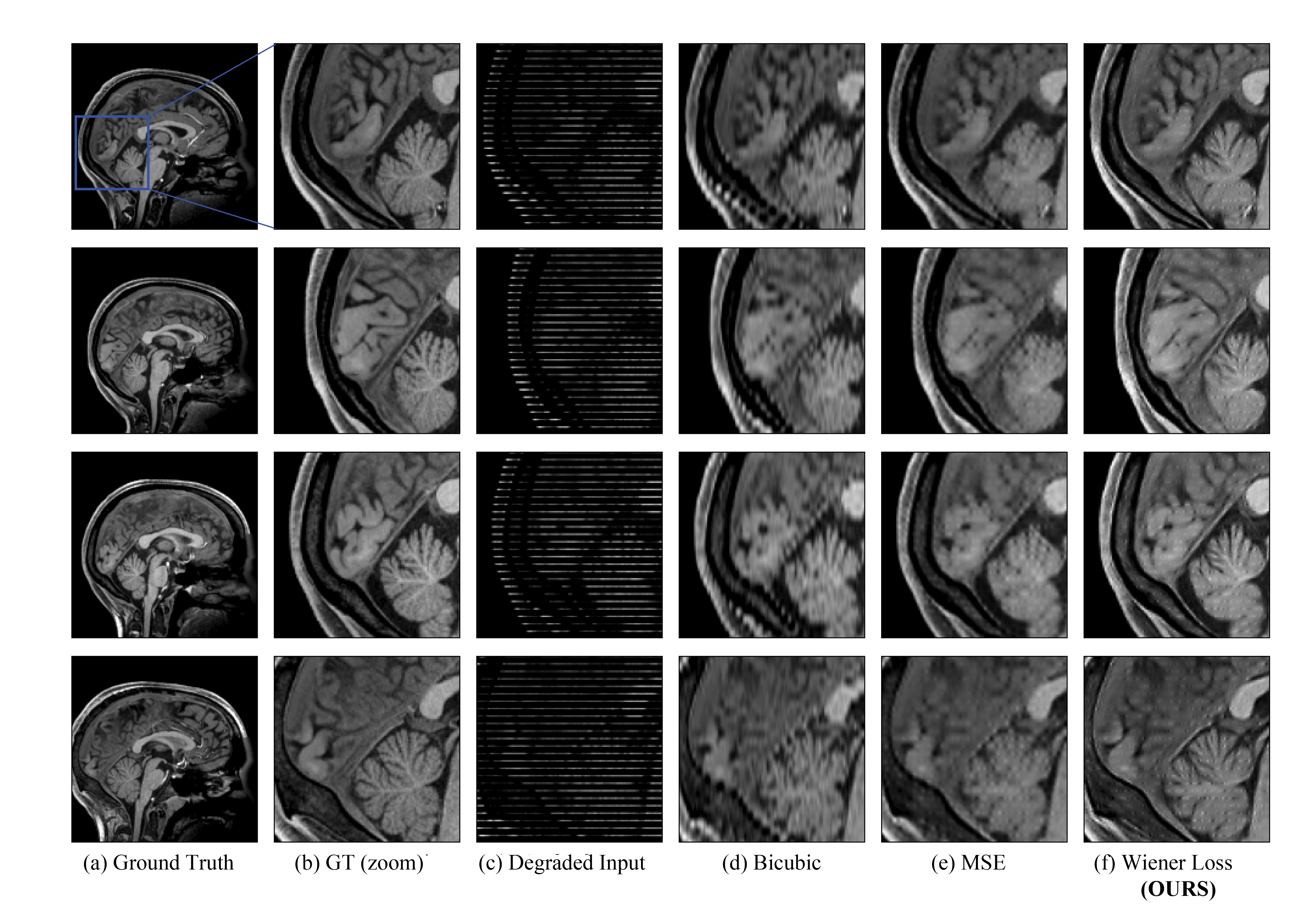}
   \caption{Ground truth (a-b) and zoomed outputs from imputing under-sampled scans (c) using bicubic interpolation (d), U-net \cite{ronneberger2015u} with MSE (e) and U-net \cite{ronneberger2015u} with Wiener loss (f). The Wiener loss is able to recover finer structures of the scan that are aliased or blurred in other methods. Results of (e-f) are obtained from the validation set.}
   \label{fig:imputation_samples}
\end{figure*}

We use the MGH-USC Human Connectome Project (HCP) Young Adult database \cite{van2012human} to test our claims. The dataset is comprised of 553 3D single-channel T1-weighted MRI scans of healthy adult brains, which we reduce to 2D by retrieving the mid-sagittal plane. The 2D scans are single-channel and of size 256 x 256. The dataset has also been de-identified with normalised anterior filtering \cite{milchenko2013obscuring} in order to remove identifying and sensitive anatomical features of patients. Our network is a U-net \cite{ronneberger2015u} composed of 3 residual units, sequenced channels of sizes 16, 32 and 64 (see supplementary materials), and \textit{Mish} \cite{misra2019mish} as activation function. Acquisition sparsity is simulated by applying a sampling mask to the full-resolution scans, serving as input to our network (Figure \ref{fig:imputation_samples}). The masked input samples have their outliers removed by capping values above the data's upper quartile range, and are scaled between 0 and 1 through a minimum-maximum value-based normalisation method. We do so to prevent statistical bias while preserving the distribution of the original data. Similarly to our previous example, we used a train/validation split of 80/20, a modified two-dimensional Laplace density function as whitening matrix $\mathbf{W}$ (see supplementary materials), and a fixed pre-whitening scalar $\lambda= 250$ for the 2D filter computation. We use a batch size of 32, and employ the Adam optimiser \cite{kingma2014adam} with a learning rate of $1e^{-2}$. We then compare the output of the model to the target scan and use this measure to train our network in a supervised setting. Results from using a standard bicubic interpolation, the MSE loss, and our Wiener loss for the U-net \cite{ronneberger2015u} imputation are presented in Figure \ref{fig:imputation_samples}. For both MSE and Wiener loss experiments, 300 epochs achieved model convergence.

In this heavily sparse problem, the results presented in Figure \ref{fig:imputation_samples} show substantial resolution improvements from using a supervised neural network to perform the imputation task. Generalisation errors have been suppressed despite the small dataset size. The use of our Wiener loss promotes imputation outputs with higher resolution of smaller features, e.g. micro structures of the cerebellum (see zoom), as well as continuity and sharpness of the skull, when compared to outputs from the MSE optimisation and bicubic interpolation. As previously mentioned, the reconstructions from the Wiener loss suffer from luminance and brightness recovery constraints, performing with poorer scores on the MAE, MSE, and SSIM metrics (Table \ref{table:metrics}). However, our Wiener loss excels at recovering perceptual quality and is better suited to restore the ground-truth distribution of the data. This is evidenced by lower scores on LPIPS \cite{zhang2018perceptual} and FID \cite{heusel2017gans} metrics. The results of our loss have increased scan resolution and contrast, which can have profound impacts during clinical analysis.

\subsection{Generative Modelling}
\label{sec:expdiffusion}

\begin{figure}[t]
  \centering
   \includegraphics[width=0.5\linewidth]{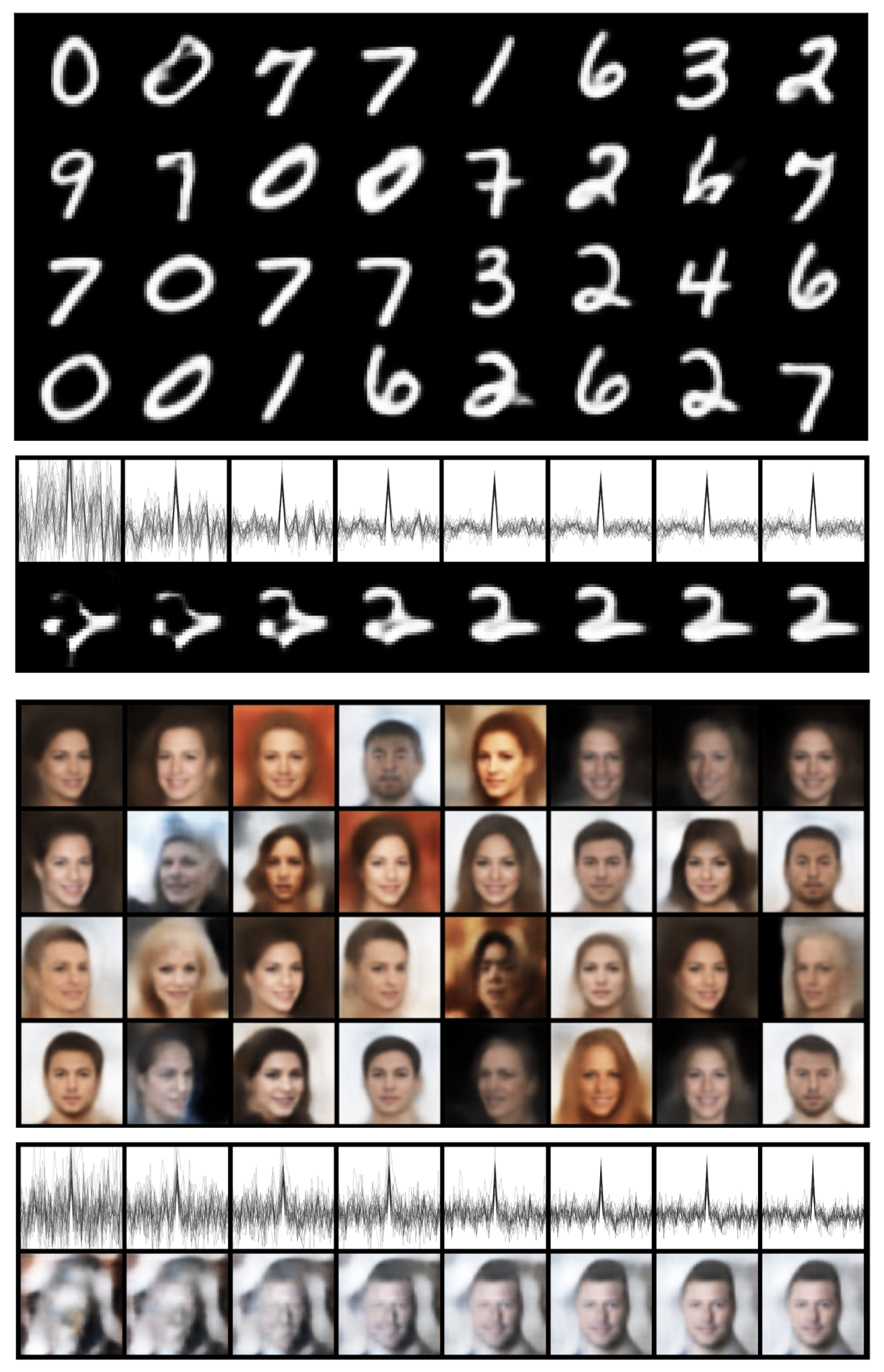}
   \caption{Latent samples generated using the Wiener diffusion, visualised using the decoder, and an illustration of the diffusion process alongside the 30 closest Wiener filters to the sample $\mathbf{x}_{t}$ (as defined by \eqref{eqn:wiener-diffusion-energy}) for MNIST (top) and CelebA (bottom).}
   \label{fig:diffusion}
\end{figure}

The third application we present is the task of image generation using the proposed Wiener diffusion. We validate this unique approach and demonstrate its behaviour by applying it to the latent space of convolutional autoencoders trained on two well known datasets, MNIST \cite{lecun-mnisthandwrittendigit-2010} and CelebA \cite{liu2015faceattributes}. Although we showcase latent space generation here, we emphasise that the Wiener diffusion can be applied directly to natural images; this will be the focus of future research. The autoencoder architectures used comprise 6 convolutional layers, 2 dense layers, and employ the Leaky ReLU activation function \cite{maas2013rectifier}, with a latent space bottleneck that is a vector of length 31 for MNIST, and 51 for CelebA. For both experiments, we use 20,000 samples to define the energy function \eqref{eqn:wiener-diffusion-energy} for the Wiener diffusion, with the initial sample $\mathbf{x}_{t=0}$ drawn from a normal distribution with a variance equivalent to the variance of the latent bottleneck. For the Wiener filter computation, we employ a pre-whitening scalar of $\lambda=1e^{-1}$, and use a modified one-dimensional Laplace density function to define the diagonal values of $\mathbf{T}$ (see supplementary materials). The MNIST diffusion is performed over $T=200$ steps, with a cosine step schedule decreasing from $\alpha_{t}=500$ to $\alpha_{t}=1$, and a cosine variance schedule increasing from $\beta_{t}=0.01$ to $\beta_{t}=0.8$. For CelebA, the diffusion is run for $T=400$ steps, with a cosine step schedule decreasing from $\alpha_{t}=3000$ to $\alpha_{t}=300$ and a cosine variance schedule increasing from $\beta_{t}=0.1$ to $\beta_{t}=4$. The generated latent samples are visualised in Figure \ref{fig:diffusion} using the corresponding decoders, alongside a visualisation demonstrating how the data-matching Wiener filters are progressively focused towards zero-lag delta functions throughout the diffusion process. 

We stress that Wiener diffusion is a non-parametric method and, although we use it on the latent space of trained autoencoders, the generative process does not require the training of a model. The benefits of this are that it is well suited to scenarios with limited data, and is also a good candidate for controllable generation as the diffusion process can be conditioned, at inference time, by specifying the data that defines the energy function \eqref{eqn:wiener-diffusion-energy}; future work will explore these hypothesized advantages.

\subsection{Image Classification}
\label{sec:expknn}

\begin{figure}[t]
  \centering
   \includegraphics[width=0.8\linewidth]{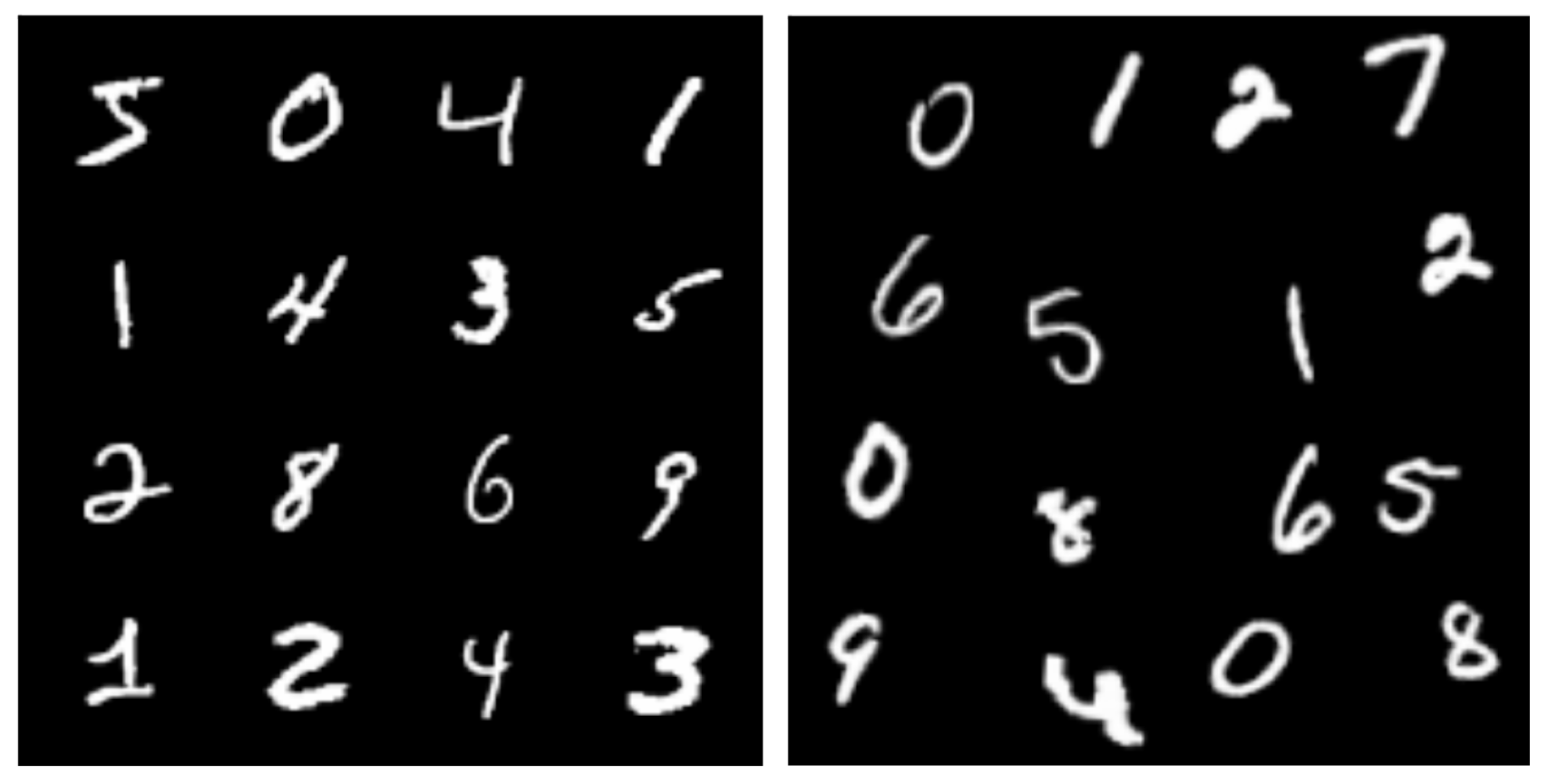}
   \caption{Image classification task to use the MNIST training data (left) to predict the affNIST testing data (right). A KNN classifier with the Manhattan distance metric achieves and accuracy of $\mathbf{14}\boldsymbol{\%}$, whereas a KNN classifier with our proposed translation invariant metric $\mathcal{L}_{\textsc{ti}}$ achieves an accuracy of $\mathbf{61}\boldsymbol{\%}$}.
   \label{fig:knn}
\end{figure}

In this final experiment, we further evidence our claim that Wiener filters provide a versatile and customisable tool for data comparison. The task we explore is the classification of digits from the affNIST dataset \cite{tieleman2013affnist}, which consist of padded MNIST images that have undergone affine transformations, using only the original (but equivalently padded) MNIST data as the training set (see Figure \ref{fig:knn}). We first evaluate a baseline model; the k-nearest neighbours (KNN) algorithm \cite{cover1967nearest}. Using a 5-fold cross-validated grid-search on the training set, the optimal configuration of the baseline KNN \cite{cover1967nearest} was obtained with the Manhattan distance metric and a k of 3. This model entirely failed to classify samples from the affNIST test set, achieving an accuracy of $\mathbf{14}\boldsymbol{\%}$. The reason for such poor performance is due to translation, an affine transformation that, from Figure \ref{fig:knn}, appears to be responsible for the predominant differences between the training and testing data distributions. 

We therefore formulate a translation invariant comparison function $\mathcal{L}_{\textsc{ti}}$, by considering the maximum value of a standardised data-matching Wiener filter as:
\begin{equation}
    \mathcal{L}_{\textsc{ti}}(\mathbf{x}, \mathbf{y})=-\textup{max}\left ( \frac{\mathbf{v}(\mathbf{x}, \mathbf{y})-\mu_{\mathbf{v}}}{\sigma_{\mathbf{v}}} \right )
\label{eqn:translation_invariant}
\end{equation}
where $\mu_{\mathbf{v}}$ is the mean value of filter $\mathbf{v}(\mathbf{x}, \mathbf{y})$, and $\sigma_{\mathbf{v}}$ is its standard deviation. If the energy of $\mathbf{v}(\mathbf{x}, \mathbf{y})$ is focused in a narrow spike, then the process of standardisation amplifies the magnitude of the spike to ensure unit standard deviation. On the other hand, if the energy of $\mathbf{v}(\mathbf{x}, \mathbf{y})$ is unfocused and distributed across all lags, then unit standard deviation is achieved without requiring a large maximum value in the standardised filter. $\mathcal{L}_{\textsc{ti}}$ is therefore only sensitive to whether the energy in the filter is focused or not; it does not care where the energy is focused in the filter and, as is illustrated in Figure \ref{fig:filters}, is thus invariant to translation. Using $\mathcal{L}_{\textsc{ti}}$ as a distance metric in KNN \cite{cover1967nearest} , with a k of 10 on the same task achieves a dramatically improved test accuracy of $\mathbf{61}\boldsymbol{\%}$. A promising direction for future research is to use the ideas discussed here to formulate optimisation metrics for deep learning where translational invariance is required.

\section{Considerations}
\label{sec:discussion}

Throughout experimentation we observe training and performance sensitivity to the choice of hyper-parameters that govern the application of our methods. These include: i) the whitening matrix $\mathbf{W}$ that controls the behaviour of the Wiener loss \eqref{eq:wiener_loss}, ii) the penalty matrix $\mathbf{T}$ that defines the shape of the energy landscape in \eqref{eqn:wiener-diffusion-energy}, and iii) the pre-whitening scalar $\lambda$ that stabilises the Wiener filter computation \eqref{eqn:fourier-wiener}. Future work will explore the systematic change in behaviour as a result of varying these hyper-parameters. For instance, using optimal whitening matrices (as discussed by \cite{kessy2018optimal}), investigating non-diagonal penalty functions, and choosing different pre-whitening values for the numerator vs the denominator which can amplify or suppress particular frequencies.

Finally, we found that the Fourier-domain implementation of the Wiener loss remains computationally competitive to when compared to the MSE loss (see Figure in supplementary materials). For MSE, the computational complexity is $O(n)$ and the memory scales as $O(n^2)$. For the Wiener loss implementation, computational complexity is $O(log \, n)$ when implemented in the Fourier domain using FFT and $O(n^2)$ in the data domain using Levinson-Durbin recursion; while memory scales as $O(n^2)$ in the Fourier domain and $O(n^4)$ in the data domain. We note here that both Fourier and data domain implementations are valid and produce equivalent results, but the later proved to be computationally intractable for most practical deep-learning problems. We also note that the current implementations are not fully optimised and that performance can be further improved.

\section{Conclusion}
\label{sec:conclusion}

Despite the immense evolution of machine learning over the last decade in terms of architecture complexity, limited progress has been made in developing optimal strategies to quantify data comparisons effectively. We address this issue by proposing a new method based on Wiener filters that promote contextual awareness in data samples. At its core lays the principle of convolutional identity as a measure of similarity between data sample pairs.

The results we present here indicate that this formulation is effective in a number of unrelated tasks. We demonstrate enhanced perceptual quality in autoencoder reconstructions, resolution improvements in imputation problems, translation invariance in classification tasks, and define a novel non-parametric generative model.

Our main contribution is a conceptual principle that holds immense potential across computer vision problems, and more general machine learning applications. So far, we have demonstrated its potential and problem-agnostic nature in four different tasks, and we are currently working towards improving our understanding of the fundamental assumptions that underlay its apparent superior performance.

\clearpage
{\small
\bibliographystyle{ieee_fullname}
\bibliography{main}
}


\end{document}